\def\year{2019}\relax
\documentclass[letterpaper]{article} 
\usepackage{aaai19}  
\usepackage{times}  
\usepackage{helvet}  
\usepackage{courier}  
\usepackage{url}  
\usepackage{graphicx}  
\usepackage{tabularx}
\usepackage{subfigure}
\usepackage{amssymb}
\usepackage{amsmath}
\usepackage{makeidx}
\usepackage{latexsym,bm}
\usepackage{verbatim}
\usepackage{booktabs}
\usepackage{algorithm,algorithmic}
\usepackage{multirow}
\begin{document}
%
\title{Aligning Domain-specific Distribution and Classifier for Cross-domain Classification from Multiple Sources}
\author{
Yongchun Zhu$^{1,2}$,
Fuzhen Zhuang$^{1,2,}$\thanks{Corresponding author: Fuzhen Zhuang} and
Deqing Wang$^{3}$
\\
$^1$Key Lab of Intelligent Information Processing of Chinese Academy of Sciences (CAS),\\
Institute of Computing Technology, CAS, Beijing 100190, China\\
$^2$University of Chinese Academy of Sciences, Beijing 100049, China \\
$^3$School of Computer Science, Beihang University, Beijing, China \\
\{zhuyongchun18s, zhuangfuzhen\}@ict.ac.cn, dqwang@buaa.edu.cn
}

\maketitle
\begin{abstract}
While Unsupervised Domain Adaptation (UDA) algorithms, i.e., there are only labeled data from source domains, have been actively studied in recent years, most algorithms and theoretical results focus on Single-source Unsupervised Domain Adaptation (SUDA). However, in the practical scenario, labeled data can be typically collected from multiple diverse sources, and they might be different not only from the target domain but also from each other. Thus, domain adapters from multiple sources should not be modeled in the same way. Recent deep learning based Multi-source Unsupervised Domain Adaptation (MUDA) algorithms focus on extracting common domain-invariant representations for all domains by aligning distribution of all pairs of source and target domains in a common feature space. However, it is often very hard to extract the same domain-invariant representations for all domains in MUDA. In addition, these methods match distributions without considering domain-specific decision boundaries between classes. To solve these problems, we propose a new framework with two alignment stages for MUDA which not only respectively aligns the distributions of each pair of source and target domains in multiple specific feature spaces, but also aligns the outputs of classifiers by utilizing the domain-specific decision boundaries. Extensive experiments demonstrate that our method can achieve remarkable results on popular benchmark datasets for image classification.
\end{abstract}

\section{Introduction}\label{sec:intro}
Recent advances in deep learning have significantly improved the state-of-the-arts across a variety of visual learning tasks~\cite{ren2015faster,he2016deep}. These achievements mainly come from the availability of large-scale labeled data for supervised learning. For a target task with the shortage of labeled data, there is a strong motivation to build effective learners that can leverage rich labeled data from a related source domain. However, due to the presence of domain shift~\cite{quionero2009dataset,pan2010survey}, the performance of the learned model might tend to degrade heavily in the target domain.

Learning a discriminative model in the presence of domain shift between training and test distributions is known as domain adaptation. In recent years, most domain adaptation algorithms focus on Single-source Unsupervised Domain Adaptation (SUDA) problem, where there are only labeled data from one single source domain. Previous SUDA methods include re-weighting the training data~\cite{jiang2007instance,huang2007correcting}, and finding a transformation in a lower-dimensional manifold that draws the source and target subspaces closer~\cite{gong2012geodesic,fernando2013unsupervised}. In recent years, most SUDA algorithms learn to map the data from both domains into a common feature space to learn domain-invariant representations by minimizing domain distribution discrepancy~\cite{long2015learning,ganin2015unsupervised,sun2016deep,long2016deep}, and the source classifier can then be directly applied to target instances. 

\begin{figure}[t!]
\centering
\begin{minipage}[b]{1\linewidth}
\centering
\includegraphics[width=.85\columnwidth]{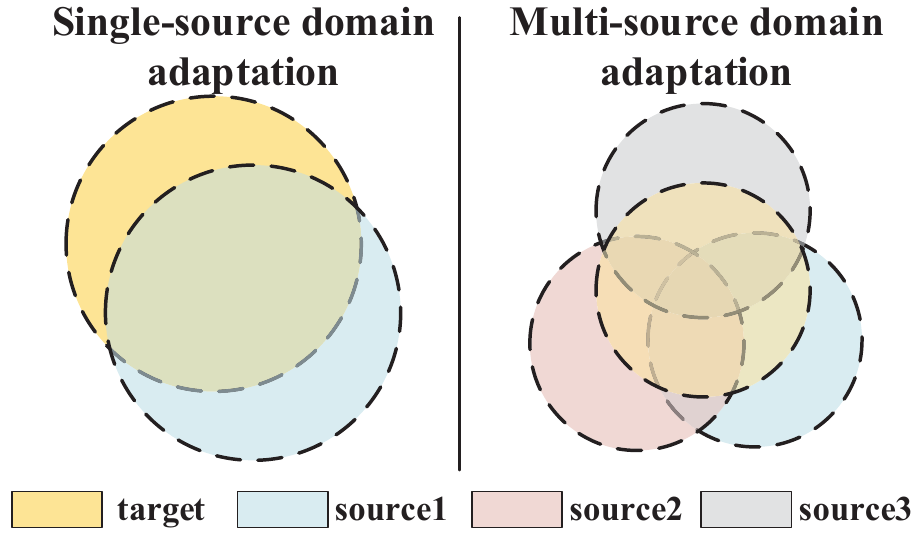}
\end{minipage}
\caption{In Single-source Unsupervised Domain Adaptation (SUDA), the distribution of source and target domains cannot be matched very well. While in Multi-source Unsupervised Domain Adaptation (MUDA), due to the shift between multiple source domains, it is much harder to match distributions of all source domains and target domains. (Best viewed in color.)}\label{UDAcmpMDA} 
\end{figure}

However, in practice, it is very likely that we have multiple source domains. Consequently, Multi-source Unsupervised Domain Adaptation (MUDA) is both feasible in practice and more valuable in performance improvement and has received considerable attention in real-world application fields~\cite{yang2007cross,duan2012domain2,jhuo2012robust,liu2016structure}. It is a common and straightforward way to combine all source domains into one single source domain and align distributions as SUDA methods do. Due to the data expansion, the methods might improve the performance. However, the improvement might not be significant, hence, it is necessary to find a better way to make full use of multiple source domains. 

Despite the rapid progress in deep learning based SUDA, a few studies have been given to deep learning based MUDA, which is much more challenging~\cite{xu2018deep}. In recent years, some works on deep learning for MUDA are proposed, and there are two common problems exist in these methods. First, they try to map all source and target domain data into a common feature space to learn common domain-invariant representations. However, it is not easy to learn domain-invariant representations even for one single source and one target domain data. As an intuitive example in Figure~\ref{UDAcmpMDA}, we can not remove the shift between one single source and one target domains, and when we try to align multiple source and target domains, the bigger mismatch degree might lead to unsatisfying performance. Second, they assume that the target domain data can be classified correctly by multiple domain-specific classifiers because they are aligned with the source domain data. However, these methods might fail to extract discriminative features because it does not consider the relationship between target samples and the domain-specific decision boundary when aligning distributions.

In this paper, we propose a new framework with two-stage alignments for MUDA to overcome both problems. The first stage is aligning the domain-specific distribution, i.e., we respectively map each pair of source and target domains data into multiple different feature spaces, and align domain-specific distributions to learn multiple domain-invariant representations. Then we train multiple domain-specific classifiers using multiple domains-invariant representations. The second stage is aligning domain-specific classifiers. The target samples near domain-specific decision boundary predicted by different classifiers might get the different labels. Hence, utilizing the domain-specific decision boundaries, we align the classifiers' output for the target samples. Extensive experiments show that our method can obtain remarkable results for MUDA on public benchmark datasets compared to the state-of-the-art methods.

The contributions of this paper are summarized as follows. (1) We propose a new two-stage alignment framework for MUDA which aligns the domain-specific distributions of each pair of source and target domains in multiple feature spaces and align the domain-specific classifiers' output for target samples. (2) We conduct comprehensive experiments on three well-known benchmarks, and the experimental results validate the effectiveness of the proposed model.

\section{Related Work}\label{sec:relatedWork} 
In this section, we will introduce the related work in two aspects: Single-source Unsupervised Domain Adaptation (SUDA) and Multi-source Unsupervised Domain Adaptation (MUDA).

\textbf{Single-source Unsupervised Domain Adaptation (SUDA)}. Recent years have witnessed many approaches to solve the visual domain adaptation problem, which is also commonly framed as the visual dataset bias problem~\cite{quionero2009dataset,pan2010survey}. Previous shallow methods for SUDA include re-weighting the training data so that they could more closely reflect those in the test distribution~\cite{jiang2007instance,huang2007correcting}, and finding a transformation in a lower-dimensional manifold that draws the source and target subspaces closer~\cite{gong2012geodesic,pan2011domain,fernando2013unsupervised}.

Some recent works bridge deep learning and domain adaptation~\cite{long2015learning,ganin2015unsupervised,tzeng2017adversarial,sun2016deep}. The two mainstreams: the one extends deep convolutional networks to domain adaptation by adding adaptation layers through which the mean embeddings of distributions are matched~\cite{tzeng2014deep,long2015learning,long2016deep}, while the other by adding a subnetwork as domain discriminator and the deep features are learned to confuse the discriminator in a domain-adversarial training paradigm~\cite{ganin2015unsupervised,tzeng2017adversarial,saito2017maximum}. And recent related work extends the adversarial methods to a generative adversarial way~\cite{bousmalis2017unsupervised,hoffman2017cycada}. 

Besides of these two mainstreams, there are diverse methods to learn domain-invariant features: DRCN~\cite{ghifary2016deep} reconstructs features to images and makes the transformed images are similar to original images. D-CORAL~\cite{sun2016deep} ``recolors" whitened source features with the covariance of features from the target domain. 

\textbf{Multi-source Unsupervised Domain Adaptation (MUDA)}. The SUDA methods mentioned above mainly consider one single source and one target domain. However, in practice, there are multiple source domains available. Due to the dataset shift among them, we can not use SUDA methods by combining all source domains into one single source domain. The research originates from A-SVM~\cite{yang2007cross} that leverages the ensemble of source-specific classifiers to tune the target categorization model, and there have been a variety of shallow models invented to tackle the MUDA problem~\cite{duan2012domain2,jhuo2012robust,liu2016structure}. MUDA also develops with theoretical supports~\cite{ben2010theory,blitzer2008learning,liu2016structure}. Blitzer et al.~\cite{blitzer2008learning} provided the first learning bound for MUDA. Mansour et al.~\cite{mansour2009domain} claimed that an ideal target hypothesis can be represented by a distribution weighted combination of source hypotheses. However, in our method, we just use the average of source hypotheses as the target hypothesis.

In recent years, some works bridge multiple source domain adaptation and deep transfer~\cite{xu2018deep,zhao2018multiple}. Xu et al.~\cite{xu2018deep} proposed to use a classifier and a domain discriminator for each pair of source and target domains, and then to vote for the target labels according to the confusion loss. Zhao et al.~\cite{zhao2018multiple} proposed to combine the gradient of multiple domain discriminators. These work focus on extracting common domain-invariant representations for all domains. However, as mentioned above, it is hard to learn common domain-invariant representations for all domains. Hence, we try to respectively map each pair of source and target domain into multiple feature spaces and extract multiple domain-invariant representations. In addition, utilizing the domain-specific decision boundaries, we align the classifiers' output for the target samples.

\begin{figure*}[ht]
\centering
\begin{minipage}[b]{1\linewidth}
\centering
\includegraphics[width=.80\columnwidth]{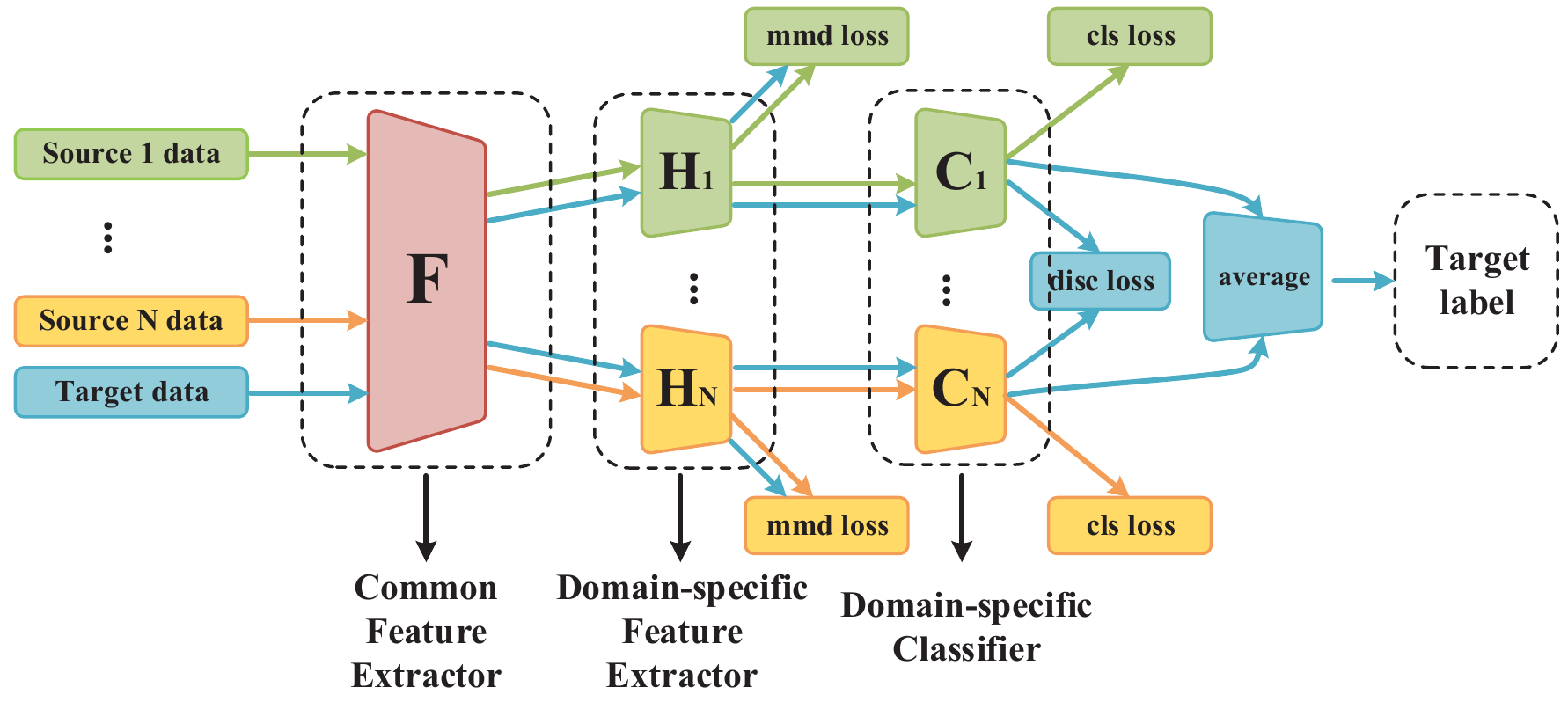}
\end{minipage}
\caption{An overview of the proposed two-stage alignment framework. Our framework receives multi-source instances with annotated ground truth and adapts to classifying the target samples. There are specific feature extractors and classifiers for each source. (Best viewed in color.)}\label{network}
\end{figure*}

\section{Method}
\label{sec:model}

In multi-source unsupervised domain adaptation, there are $N$ different underlying source distributions denoted as $\{p_{sj}(x,y)\}^N_{j=1}$, and the labeled source domain data $\{(X_{sj},Y_{sj})\}^N_{j=1}$ are drawn from these distributions respectively, where $X_{sj}=\{x_i^{sj}\}_{i=1}^{|X_{sj}|}$ represents samples from source domain $j$ and $Y_{sj}=\{y_i^{sj}\}_{i=1}^{|X_{sj}|}$ is the corresponding ground-truth labels. Also, we have target distribution $p_t(x,y)$, from which target domain data $X_t=\{x_i^t\}_{i=1}^{|X_t|}$ are sampled yet without label observation $Y_t$. 

In recent years, some works bridge deep learning and multi-source domain adaptation~\cite{xu2018deep,zhao2018multiple}, and they minimize a distance loss between each pair of source and target domains to learn common domain-invariant representations in a common feature space for all domains. The formal representation can be:
\begin{equation}
\begin{split}
\min_{F,C} &\sum^{N}_{j=1} \mathbf{E}_{x\sim X_{sj}} J(C(F( \mathbf{x}^{sj}_i )), \mathbf{y}^{sj}_i)\\
&+ \lambda \sum_{j=1}^{N} \hat{D}(F(X_{sj}), F(X_t)),
\label{preeq}
\end{split}
\end{equation}
where $J(\cdot,\cdot)$ is the cross-entropy loss function (classification loss) and $\hat{D}(\cdot,\cdot)$ is an estimate of the discrepancy between two domains, such as MMD~\cite{gretton2012kernel,long2015learning}, CORAL~\cite{sun2016deep}, Confusion loss~\cite{ganin2015unsupervised,tzeng2015simultaneous}. $F(\cdot)$ is the feature extractor to map all domains into a common feature space, and $C(\cdot)$ is the classifier. The common problem with these methods is that they mainly focus on learning common domain-invariant representations for all domains and do not consider domain-specific decision boundaries between classes. However, it is not an easy task. Actually, extracting domain-invariant representations for each pair of source and target domains respectively is easier than extracting common domain-invariant representations for all domains. In addition, the target samples near domain-specific decision boundary predicted by different classifiers might get the different labels. Hence, utilizing the domain-specific decision boundaries, we align the classifiers' outputs for the target samples. Therefore, we propose a new two-stage alignment framework to overcome these problems.

First alignment stage is aligning the domain-specific distributions for each pair of source and target domains. The way to extract multiple domain-invariant representations for each pair of source and target domain is that mapping each of them into specific feature spaces and matching their distributions. To map each pair of source and target domains into a specific feature space, the easiest way is to train multiple networks. However, this would spend a lot of time. Hence, we propose to divide the network into two part. Specifically, the first part shares a subnetwork to learn some common features for all domains, and the second part contains $N$ domain-specific subnetworks that do not share the weights with each other for each pair of source and target domains. For each unshared subnetwork, we learn a domain-specific classifier. However, the target samples near domain-specific decision boundary predicted by different classifiers might get the different labels. Hence, utilizing the domain-specific decision boundaries, the second alignment stage is aligning the domain-specific classifiers' output for the target samples. In paper~\cite{xu2018deep}, they proposed a complex voting method for multiple classifiers, in our method the complex voting method is not needed due to the second stage alignment. 

\subsection{Two-stage alignment Framework}
Our framework consists of three components, i.e., a common feature extractor, domain-specific feature extractors, domain-specific classifiers, as shown in Figure~\ref{network}.

\textbf{Common feature extractor} We propose a common subnetwork $f(\cdot)$ to extract common representations for all domains, which map the images from the original feature space into a common feature space.

\textbf{Domain-specific feature extractor} We want each pair of source and target domain data could be mapped into a specific feature space. Given a batch images $x^{sj}$ from source domain $(X_{sj}, Y_{sj})$ and a batch images $x^t$ from target domain $X_t$, these domain-specific feature extractors receive the common features $f(x^{sj})$ and $f(x^t)$ from common feature extractor. Then, there are $N$ unshared domain-specific subnetworks $h_j(\cdot)$ for each source domain $(X_{sj}, Y_{sj})$, which map each pair of source and target domains into a specific feature space. 

The aim of deep domain adaptation is to learn domain-invariant representations, and there are several methods to achieve this goal in recent years, such as mmd loss~\cite{gretton2012kernel,long2015learning}, adversarial loss~\cite{ganin2015unsupervised,tzeng2015simultaneous}, coral loss~\cite{sun2016deep}, reconstruction loss~\cite{ghifary2016deep}. Here we choose the MMD method to reduce the distribution discrepancy between domains. 

\textbf{Domain-specific classifier} $C$ is a multi-output net composed by $N$ domain-specific predictor $\{C_{j}\}^N_{j=1}$. Each predictor $C_{j}$ is a softmax classifier, and receives the specific domain-invariant feature after domain-specific feature extractor $H(F(x))$ for $j$-th source domain. For each classifier, we add a classification loss using cross entropy, which is formulated as:
\begin{equation}
\mathcal{L}_{cls} = \sum^{N}_{j=1} \mathbf{E}_{x\sim X_{sj}} J(C_{j}(H_{j}(F( \mathbf{x}^{sj}_i) )), \mathbf{y}^{sj}_i).
\label{classification}
\end{equation}

\subsection{Domain-specific Distribution Alignment}
To achieve the first alignment stage (align distributions for each pair of source and target domains), we choose Maximum Mean Discrepancy (MMD)~\cite{gretton2012kernel} as our estimate of the discrepancy between two domains. MMD is a kernel two-sample test which rejects or accepts the null hypothesis $p=q$ based on the observed samples. The basic idea behind MMD is that if the generating distributions are identical, all the statistics are the same. Formally, MMD defines the following difference measure:
\begin{equation}
D_\mathcal{H}(p,q) \triangleq \| \mathbf{E}_p[\phi (\mathbf{x}^s)]-\mathbf{E}_q[\phi (\mathbf{x}^t)]\|^2_\mathcal{H},
\end{equation}
\label{mmd}
where $\mathcal{H}$ is the reproducing kernel Hillbert space (RKHS) endowed with a characteristic kernel $k$. Here $\phi(\cdot)$ denotes some feature map to map the original samples to RKHS and the kernel $k$ means $k(\mathbf{x}^s,\mathbf{x}^t)=\langle \phi (\mathbf{x}^s), \phi (\mathbf{x}^t) \rangle$ where $\langle \cdot , \cdot \rangle$ represents inner product of vectors. The main theoretical result is that $p=q$ if and only if $D_\mathcal{H}(p,q)=0$~\cite{gretton2012kernel}. In practice, an estimate of the MMD compares the square distance between the empirical kernel mean embeddings as
\begin{equation}
\begin{split}
\hat{D}_\mathcal{H}(p,q)&=\left\| \frac{1}{n_s} \sum_{\mathbf{x}_i\in \mathcal{D}_s }\phi (\mathbf{x}_i)-\frac{1}{n_t} \sum_{\mathbf{x}_j\in \mathcal{D}_t}\phi (\mathbf{x}_j)\right\|^2_\mathcal{H},
\label{unbiased-mmd}
\end{split}
\end{equation}
where $\hat{D}_\mathcal{H}(p,q)$ is an unbiased estimator of $D_\mathcal{H}(p,q)$. We use Equation~(\ref{unbiased-mmd}) as the estimate of the discrepancy between each source domain and target domain. The MMD loss is reformulated as:
\begin{equation}
\mathcal{L}_{mmd} = \frac{1}{N} \sum^{N}_{j=1} \hat{D}(H_{j}(F(X_{sj})), H_{j}(F(X_t))),
\label{mmdloss}
\end{equation}
Each specific feature extractor could learn domain-invariant representations for each pair of source and target domain by minimizing the Equation~\ref{mmdloss}.

\subsection{Domain-specific Classifier Alignment}
The target samples near the class boundaries are more likely to be misclassified by the classifiers learned from source samples. The classifiers are trained on different source domains, hence they might have the disagreement on the prediction for target samples especially the target samples near class boundaries. Intuitively, the same target sample predicted by different classifiers should get the same prediction. Hence, the second alignment stage is to minimize the discrepancy among all classifiers. In this paper, we utilize the absolute values of the difference between all pairs of classifiers' probabilistic outputs of target domain data as discrepancy loss:
\begin{equation}
\begin{split}
\mathcal{L}_{disc} = &\frac{2}{N\times(N-1)}\sum^{N-1}_{j=1} \sum^{N}_{i=j+1} \mathbf{E}_{x\sim X_{t}}[| C_i( H_i( F(x_k) ) )\\
& - C_j( H_j( F(x_k) ) ) |],
\label{discloss}
\end{split}
\end{equation}
In ~\cite{xu2018deep}, they propose a target classification operator to combine the multiple source classifiers. However, it is complex to vote the label for target samples. By minimizing the Equation~(\ref{discloss}), the probabilistic outputs of all classifiers are similar. Finally, to predict the labels of target samples, we compute the average of all classifier outputs.

\begin{algorithm} [h]
    \caption{Multiple Feature Spaces Adaptation Network (MFSAN)}\label{alg}
    \begin{algorithmic}[1]
    \STATE{Give the number of training iterations $T$}
    \FOR{$t$ in 1 : $T$}
        \STATE {Randomly sample $m$ images $\{ x_i^{sj}, y_i^{sj} \}_{i=1}^m$ from one of $N$ source domains  $\{ (X_{sj}, Y_{sj}) \}_{j=1}^N$.}
        \STATE {Sample $m$ images $\{ x_i^t \}_{i=1}^m$ from target domain ${(X_t)}$.}
        \STATE {Feed source and target samples to common feature extractor to get the common latent representations $F(x_i^{sj})$ and $F(x_i^{t})$.}
        \STATE {Feed common latent representations of source samples to domain-specific feature extractor to get domain-specific representations of source samples $H_j(F(x_i^{sj}))$.}
        \STATE {Feed domain-specific representations of source samples $H_j(F(x_i^{sj}))$ to domain-specific classifier to get $C_j( H_j(F(x_i^{sj})) )$, and the classification is computed as Equation~(\ref{classification}).}
        \STATE {Feed common latent representation of target samples to all domain-specific extractor to get domain-specific representations of target samples $H_1(F(x_i^{t1})), \cdots, H_N(F(x_i^{tN}))$,}
        \STATE {Use $H_j(F(x_i^{sj}))$ and $H_j(F(x_i^{t1}))$ to calculate mmd loss as Equation~(\ref{mmdloss}).}
        \STATE {Use $H_1(F(x_i^{t1})), \cdots, H_N(F(x_i^{tN}))$ to compute the disc loss as Equation~(\ref{discloss}).}
        \STATE {Update the common feature extractor $F$, multiple domain-specific feature extractor $H_1,\cdots, H_N$ and multiple classifier $C_1,\cdots,C_N$ by minimizing the total loss in Equation~(\ref{totalloss}).}
    \ENDFOR
    \end{algorithmic}
\end{algorithm}

\subsection{Multiple Feature Spaces Adaptation Network}
Learning common domain-invariant representations is difficult for multiple source domains. In addition, The target samples near the class boundaries are likely to be misclassified. To this end, we propose a Multiple Feature Spaces Adaptation Network (MFSAN for short). Specifically, this network includes two alignment stages, which are to learn source specific domain-invariant representations and align the classifiers' output for target samples. Our framework is composed by a common feature extractor, $N$ domain-specific feature extractors, and $N$ source specific classifiers. Overall, the loss of our method is consist of three parts, classification loss, mmd loss, disc loss. For details, by minimizing classification loss, the network could accurately classify the source domain data; by minimizing mmd loss to learn domain-invariant representations; by minimizing disc loss to reduce the discrepancy among classifiers. The total loss is formulated as,
\begin{equation}
\begin{split}
\mathcal{L}_{total} = \mathcal{L}_{cls} + \lambda \mathcal{L}_{mmd} + \gamma \mathcal{L}_{disc}.
\label{totalloss}
\end{split}
\end{equation}
Since training deep CNNs requires a large amount of labeled data that is prohibitive for many domain adaptation applications, we start with the CNN models pre-trained on ImageNet 2012 data and fine-tune it as ~\cite{long2016deep}. The training mainly follows standard mini-batch stochastic gradient descent (SGD) algorithm. Our method is a general framework for Multi-source Unsupervised Domain Adaptation(MUDA). The $\mathcal{L}_{mmd}$ could be replaced by other adaptation methods, such as adversarial loss, coral loss. And the $\mathcal{L}_{disc}$ could be replaced by other loss, such as L2 regularization. The whole procedure is summarized in Algorithm~\ref{alg}.

\section{Experiments}\label{experiment}
We evaluate the Multiple Feature Spaces Adaptation Network (MFSAN) against state-of-the-art domain adaptation methods on three datasets: \textbf{ImageCLEF-DA}, \textbf{Office-31} and \textbf{Office-Home}. Our code will be available at: \url{https://github.com/easezyc/deep-transfer-learning}

\subsection{Data Preparation}

\textbf{ImageCLEF-DA}\footnote{http://imageclef.org/2014/adaptation.} is a benchmark dataset for ImageCLEF 2014 domain adaptation challenge, which is organized by selecting the 12 common categories shared by the following three public datasets, each is considered as a domain: \emph{Caltech-256} (\textbf{C}), \emph{ImageNet ILSVRC 2012} (\textbf{I}), and \emph{Pascal VOC 2012} (\textbf{P}). There are 50 images in each category and 600 images in each domain. We use all domain combinations and build three transfer tasks: \textbf{I}, \textbf{C} $\rightarrow$ \textbf{P}; \textbf{I}, \textbf{P} $\rightarrow$ \textbf{C}; \textbf{C}, \textbf{P} $\rightarrow$ \textbf{I}.

\textbf{Office-31}~\cite{saenko2010adapting} is a benchmark for domain adaptation, comprising 4,110 images in 31 classes collected from three distinct domains: \emph{Amazon}(\textbf{A}), which contains images downloaded from amazon.com, \emph{Webcam}(\textbf{W}) and \emph{DSLR}(\textbf{D}), which contain images taken by web camera and digital SLR camera with different photographical settings. The images in each domain are unbalanced. The images in each domain are unbalanced. To enable unbiased evaluation, we evaluate all methods on all three transfer tasks \textbf{A}, \textbf{W} $\rightarrow$ \textbf{D}; \textbf{A}, \textbf{D} $\rightarrow$ \textbf{W}; \textbf{D}, \textbf{W} $\rightarrow$ \textbf{A}.

\textbf{Office-Home}~\cite{venkateswara2017deep} is a new dataset which consists 15,588 images larger than Office-31 and ImageCLEF-DA. It consists of images from 4 different domains: Artistic images (\textbf{A}), Clip Art (\textbf{C}), Product images (\textbf{P}) and Real-World images (\textbf{R}). For each domain, the dataset contains images of 65 object categories collected in office and home settings. We use all domain combinations and build four transfer tasks: \textbf{C}, \textbf{P}, \textbf{R} $\rightarrow$ \textbf{A}; \textbf{A}, \textbf{P}, \textbf{R} $\rightarrow$ \textbf{C}; \textbf{A}, \textbf{C}, \textbf{R} $\rightarrow$ \textbf{P}; \textbf{A}, \textbf{C}, \textbf{P} $\rightarrow$ \textbf{R}.

\subsection{Baselines and Implementation Details}\label{implementation}
\textbf{Baselines}
There is a small amount of MUDA work on real-world visual recognition benchmarks. In our experiment, we introduce a recent deep MUDA method Deep Cocktail Network (\textbf{DCTN})~\cite{xu2018deep} as the multi-source baselines. Besides, We compare MFSAN with various kinds of SUDA methods, including Deep Convolutional Neural Network \textbf{ResNet}~\cite{he2016deep}, Deep Domain Confusion (\textbf{DDC})~\cite{tzeng2014deep}, Deep Adaptation Network (\textbf{DAN})~\cite{long2015learning}, Deep CORAL (\textbf{D-CORAL})~\cite{sun2016deep}, Reverse Gradient (\textbf{RevGrad})~\cite{ganin2015unsupervised} and Residual Transfer Network (\textbf{RTN})~\cite{long2016unsupervised}. Since those methods perform in single-source setting, we introduce three MUDA standards for different purposes: (1) Source combine: all source domains are combined together into a traditional single-source v.s. target setting. (2) Single best: among the multiple source domains, we report the best single source transfer results. (3) Multi-source: the results of 
MUDA methods. The first standard testifies whether the multiple sources are valuable to exploit; the second standard evaluates whether we can further improve the best SUDA via introducing other sources; the third demonstrates the effectiveness of our MFSAN.

To further validate the effectiveness of mmd loss and diff loss, we also evaluate several variants of MFSAN: (1)~\textbf{MFSAN$_{disc}$} , without considering the mmd loss; (2)~\textbf{MFSAN$_{mmd}$}, without considering the disc loss; (3)~\textbf{MFSAN}, considering both disc loss and mmd loss. For all domain-specific feature extractors, we use the same structure (conv(1x1), conv(3x3), conv(1x1)), and at the end of the network, we reduce the channels to 256 like DDC~\cite{tzeng2014deep}.

\begin{table}[!th]
\small
\centering
\caption{Performance Comparison of Classification Accuracy (\%) on Office-31 Dataset.}
\label{tab:MDAoffice31}
\begin{tabular}{@{}p{1cm}<{\centering}p{1.5cm}<{\centering}p{1cm}<{\centering}p{1cm}<{\centering}p{1cm}<{\centering}p{0.8cm}<{\centering}@{}}
\toprule
Standards & Method & A,W $\rightarrow$ D & A,D $\rightarrow$ W & D,W $\rightarrow$ A & Avg \\
\midrule
& ResNet & 99.3 & 96.7 & 62.5 & 86.2\\
& DDC & 98.2 & 95.0 & 67.4 & 86.9\\
Single & DAN & 99.5 & 96.8 & 66.7 & 87.7\\
Best & D-CORAL & 99.7 & 98.0 & 65.3 & 87.7\\
& RevGrad & 99.1 & 96.9 & 68.2 & 88.1\\
& RTN & 99.4 & 96.8 & 66.2 & 87.5\\
\midrule
& DAN & 99.6 & 97.8 & 67.6 & 88.3\\
Source & D-CORAL & 99.3 & 98.0 & 67.1 & 88.1\\
Combine & RevGrad & 99.7 & 98.1 & 67.6 & 88.5\\
\midrule
& DCTN & 99.3 & 98.2 & 64.2 & 87.2\\
Multi- & MFSAN$_{disc}$ & 99.7 & 97.9 & 68.1 & 88.6\\
Source & MFSAN$_{mmd}$ & \textbf{99.9} & 98.3 & 71.5 & 89.9\\
& MFSAN & 99.5 & \textbf{98.5} & \textbf{72.7} & \textbf{90.2}\\
\bottomrule
\end{tabular}
\end{table}

\begin{table}[!th]
\small
\centering
\caption{Performance Comparison of Classification Accuracy (\%) on Image-CLEF Dataset.}
\label{tab:MDAImage-CLEF}
\begin{tabular}{@{}p{1cm}<{\centering}p{1.5cm}<{\centering}cccp{1cm}<{\centering}@{}}
\toprule
Standards & Method & I,C $\rightarrow$ P & I,P $\rightarrow$ C & P,C $\rightarrow$ I & Avg \\
\midrule
& ResNet & 74.8 & 91.5 & 83.9 & 83.4\\
& DDC & 74.6 & 91.1 & 85.7 &  83.8\\
Single & DAN & 75.0 & 93.3 & 86.2 & 84.8\\
Best & D-CORAL & 76.9 & 93.6 & 88.5 & 86.3\\
& RevGrad & 75.0 & \textbf{96.2} & 87.0 & 86.1\\
 & RTN & 75.6 & 95.3 & 86.9 & 85.9\\
\midrule
& DAN & 77.6 & 93.3 & 92.2 & 87.7\\
Source  & D-CORAL & 77.1 & 93.6 & 91.7 & 87.5\\
Combine
 & RevGrad & 77.9 & 93.7 & 91.8 & 87.8\\
\midrule
& DCTN & 75.0 & 95.7 & 90.3 & 87.0\\
Multi- & MFSAN$_{disc}$ & 78.0 & 95.0 & 92.5 & 88.5\\
Source & MFSAN$_{mmd}$ & 78.7 & 94.8 & 93.1 & 88.9 \\
& MFSAN & \textbf{79.1} & 95.4 & \textbf{93.6} & \textbf{89.4}\\
\bottomrule
\end{tabular}
\end{table}

\begin{table}[!th]
\small
\centering
\caption{Performance Comparison of Classification Accuracy (\%) on Office-Home Dataset.} \label{tab:MDAOfficeHome}
\begin{tabular}{@{}p{1cm}<{\centering}p{1.5cm}<{\centering}p{0.7cm}<{\centering}p{0.7cm}<{\centering}p{0.7cm}<{\centering}p{0.7cm}<{\centering}c@{}}
\toprule
\multirow{2}{*}{Standards} & \multirow{2}{*}{Method} & C,P,R  & A,P,R & A,C,R & A,C,P  & \multirow{2}{*}{Avg} \\
 & & $\rightarrow$ A &  $\rightarrow$ C &  $\rightarrow$ P & $\rightarrow$ R &  \\
\midrule
& ResNet & 65.3 & 49.6 & 79.7 & 75.4 & 67.5\\
Single & DDC & 64.1 & 50.8 & 78.2 & 75.0 & 67.0\\
Best 
& DAN & 68.2 & 56.5 & 80.3 & 75.9 & 70.2\\
& D-CORAL & 67.0 & 53.6 & 80.3 & 76.3 & 69.3\\
& RevGrad & 67.9 & 55.9 & 80.4 & 75.8 & 70.0\\
\midrule
& DAN & 68.5 & 59.4 & 79.0 & 82.5 & 72.4\\
Source 
& D-CORAL & 68.1 & 58.6 & 79.5 & \textbf{82.7} & 72.2\\
Combine & RevGrad & 68.4 & 59.1 & 79.5 & \textbf{82.7} & 72.4\\
\midrule
& MFSAN$_{disc}$ & 69.8 & 60.2 & 80.2 & 81.0 & 72.8\\
Multi- & MFSAN$_{mmd}$ & 71.1 & 61.9 & 79.3 & 80.8 & 73.3\\
Source & MFSAN & \textbf{72.1} & \textbf{62.0} & \textbf{80.3} & 81.8 & \textbf{74.1}\\
\bottomrule
\end{tabular}
\end{table}

\begin{table}[!th]
\small
\centering
\caption{Classification Accuracy (\%) on Office-31 Dataset for
MFSAN with and without disc Loss.} \label{tab:diffoffice31}
\begin{tabular}{@{}p{1.5cm}<{\centering}p{1.5cm}<{\centering}p{0.8cm}<{\centering}p{0.8cm}<{\centering}p{0.8cm}<{\centering}p{0.8cm}<{\centering}@{}}
\toprule
Standards & Method & A,W $\rightarrow$ D & A,D $\rightarrow$ W & D,W $\rightarrow$ A & Avg \\
\midrule
 & S1 & 97.7 & 95.0 & 68.3 & 87.0\\
MFSAN$_{mmd}$ & S2 & 85.5 & 89.0 & 71.0 & 81.8\\
 & Avg & \textbf{99.9} & 98.3 & 71.5 & 89.9\\
\midrule
 & S1 & 97.3 & 97.6 & 72.5 & 89.1\\
MFSAN & S2 & 96.6 & 97.7 & 72.4 & 88.9\\
 & Avg & 99.5 & \textbf{98.5} & \textbf{72.7} & \textbf{90.2}\\
\bottomrule
\end{tabular}
\end{table}

\begin{figure*}[!th]
	\centering
	\subfigure[DAN (Single Source): \textbf{D, A}]{
		\begin{minipage}[b]{0.18\linewidth}
			\centering
			\includegraphics[width=.95\columnwidth,height=.95\columnwidth]{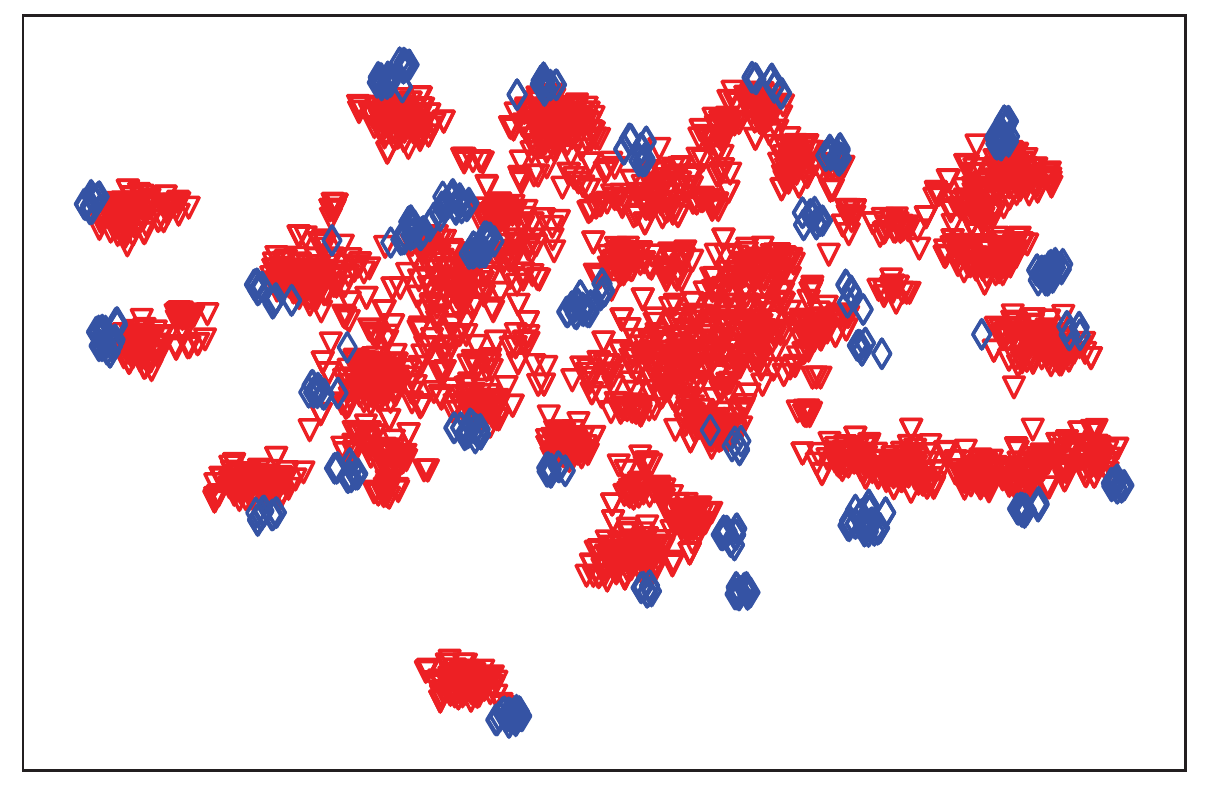}
			\label{fig:5a}
		\end{minipage}
	}
	\subfigure[DAN (Source Combine): \textbf{D, A}]{
		\begin{minipage}[b]{0.18\linewidth}
			\centering
			\includegraphics[width=.95\columnwidth,height=.95\columnwidth]{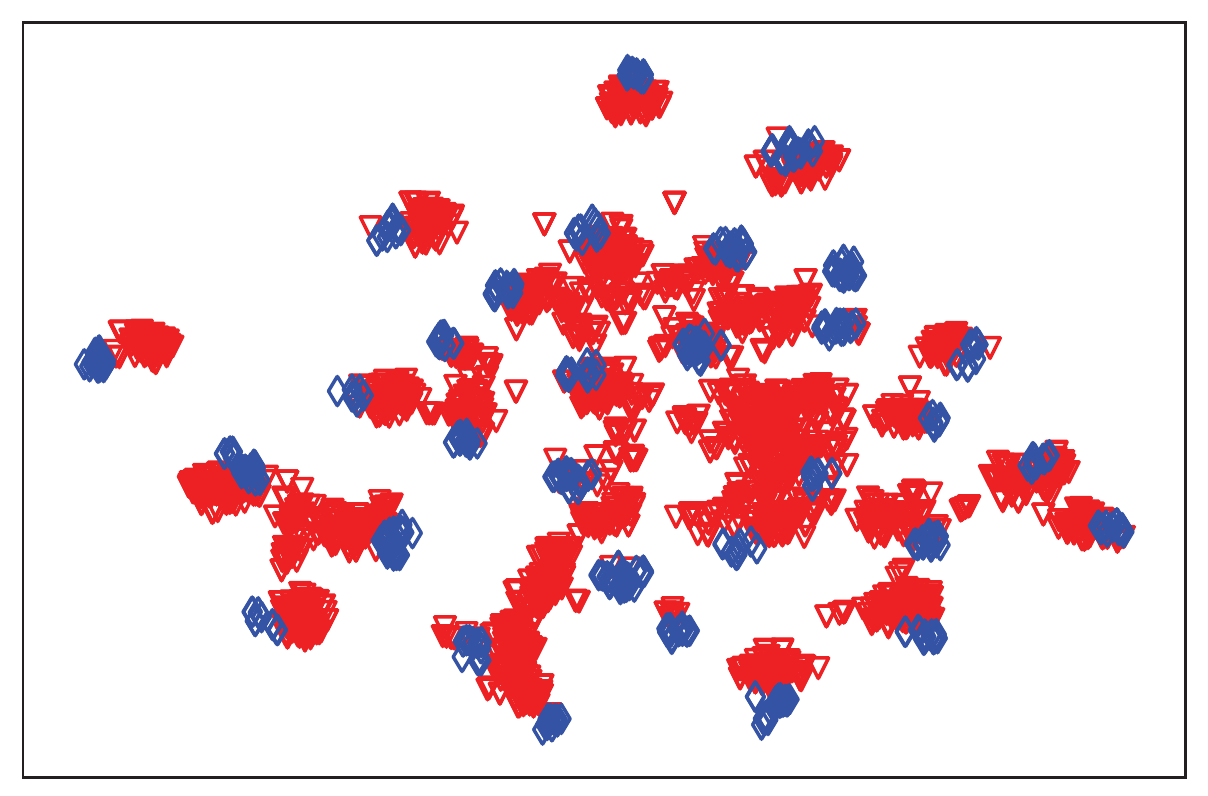}
			\label{fig:4a}
		\end{minipage}
	}
	\subfigure[DAN (Source Combine): \textbf{W, A}]{
		\begin{minipage}[b]{0.18\linewidth}
			\centering
			\includegraphics[width=.95\columnwidth,height=.95\columnwidth]{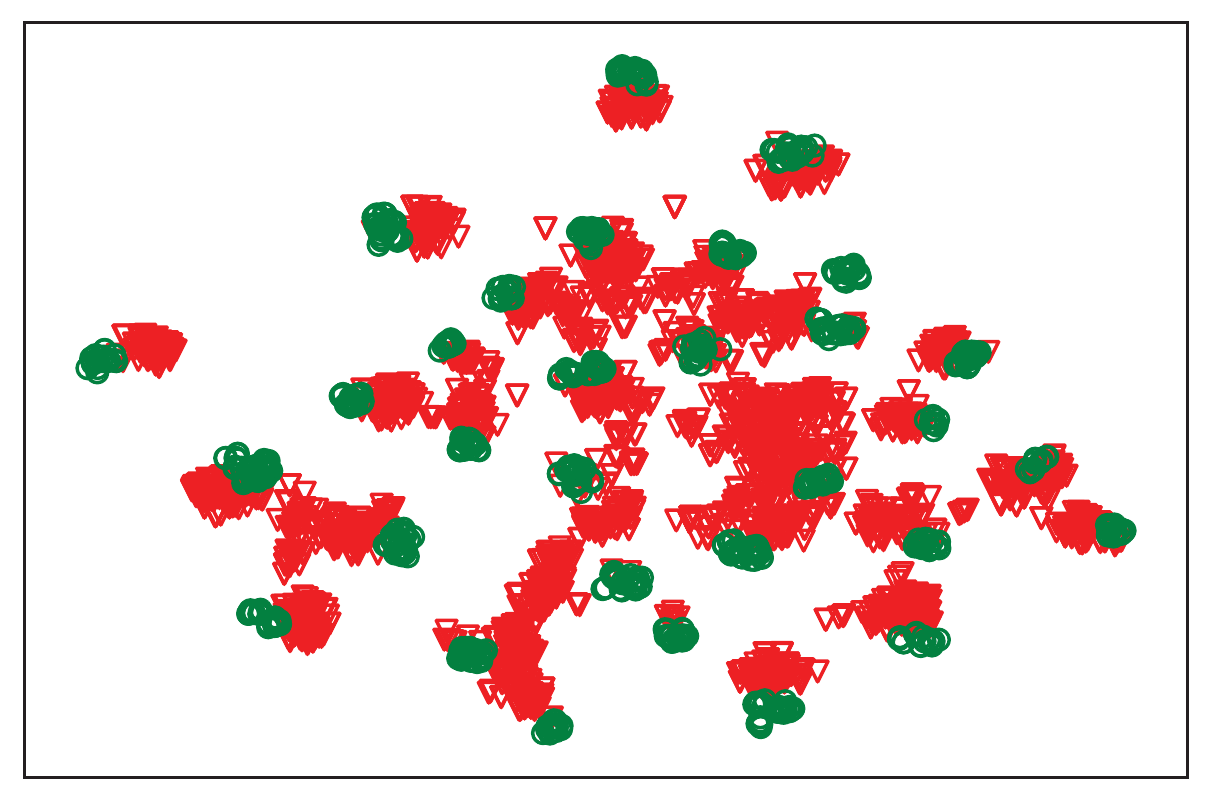}
			\label{fig:4b}
		\end{minipage}
	}
	\subfigure[MFSAN: \textbf{D, A}]{
		\begin{minipage}[b]{0.18\linewidth}
			\centering
			\includegraphics[width=.95\columnwidth,height=.95\columnwidth]{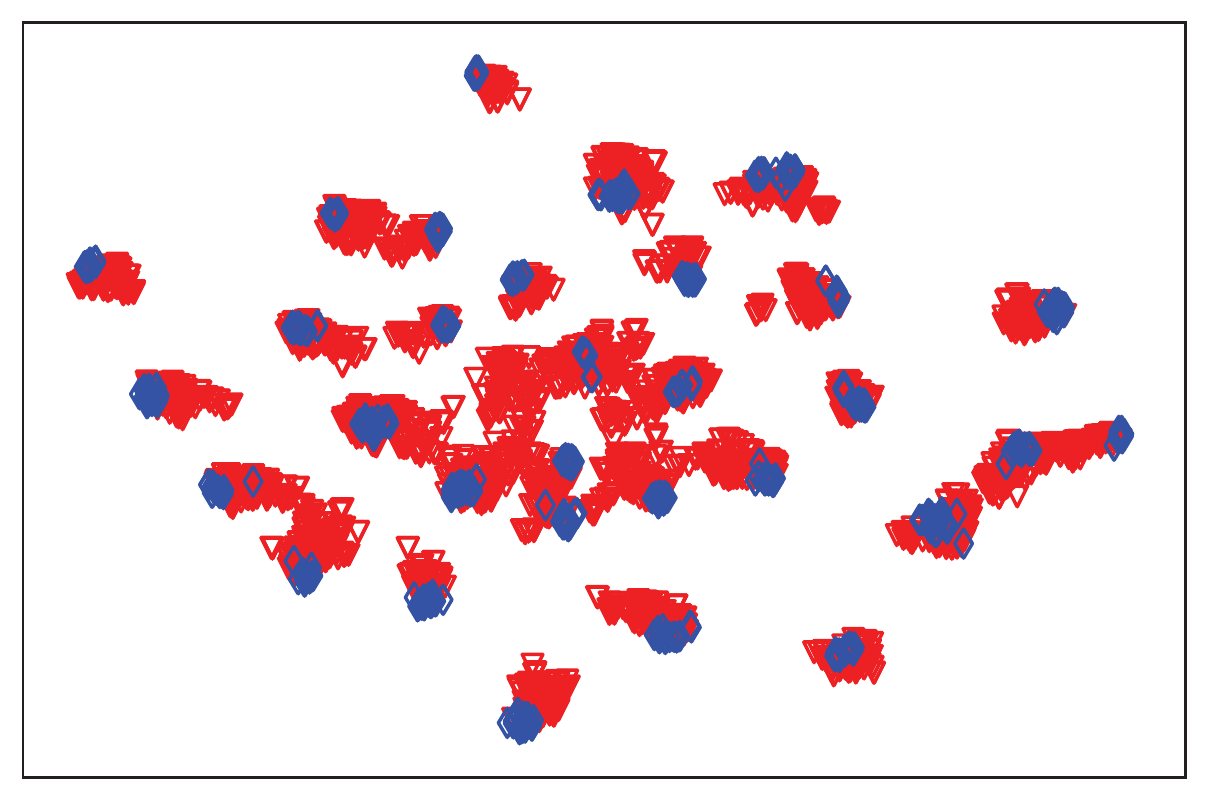}
			\label{fig:4c}
		\end{minipage}
	}
	\subfigure[MFSAN: \textbf{W, A}]{
		\begin{minipage}[b]{0.18\linewidth}
			\centering
			\includegraphics[width=.95\columnwidth,height=.95\columnwidth]{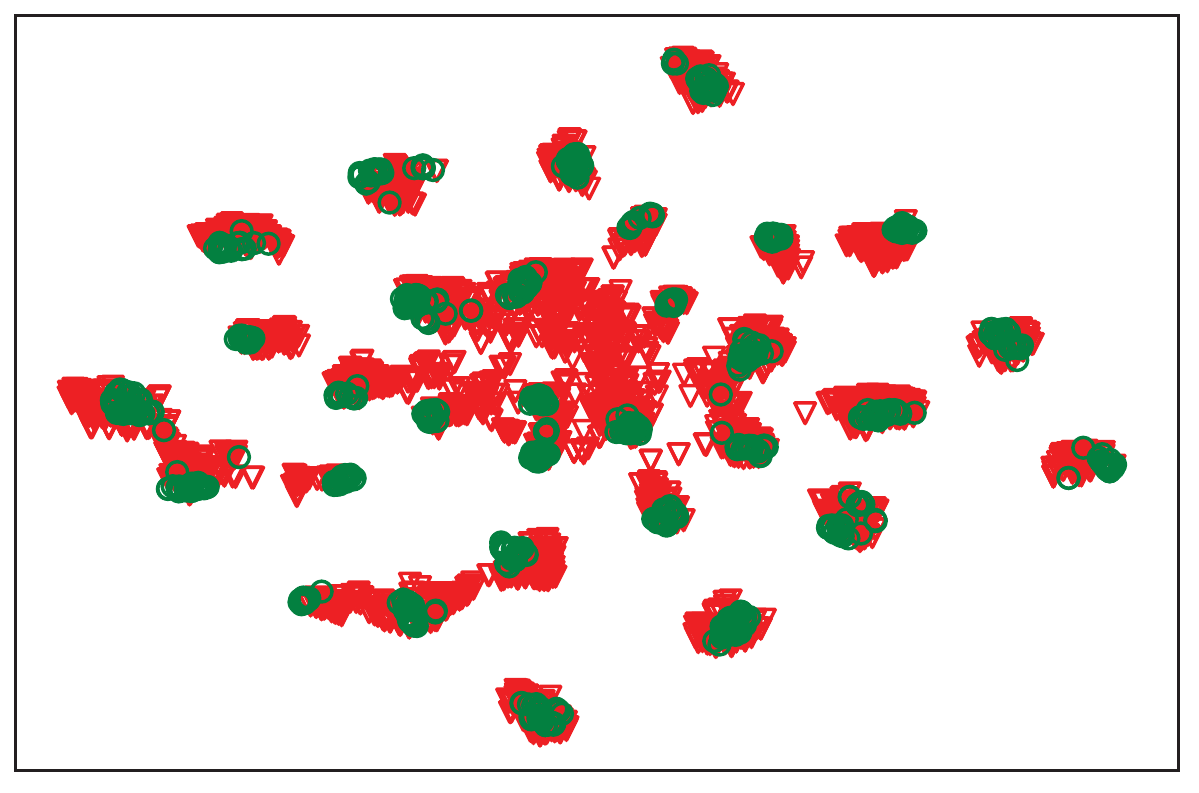}
			\label{fig:4d}
		\end{minipage}
	}
	\caption{The Visualization of Latent Representations of Source and Target Domains.}\label{fig:4}
\end{figure*}

\begin{figure}[!th]
	\centering
	\subfigure[Convergence]{
		\centering
		\includegraphics[width=.47\columnwidth]{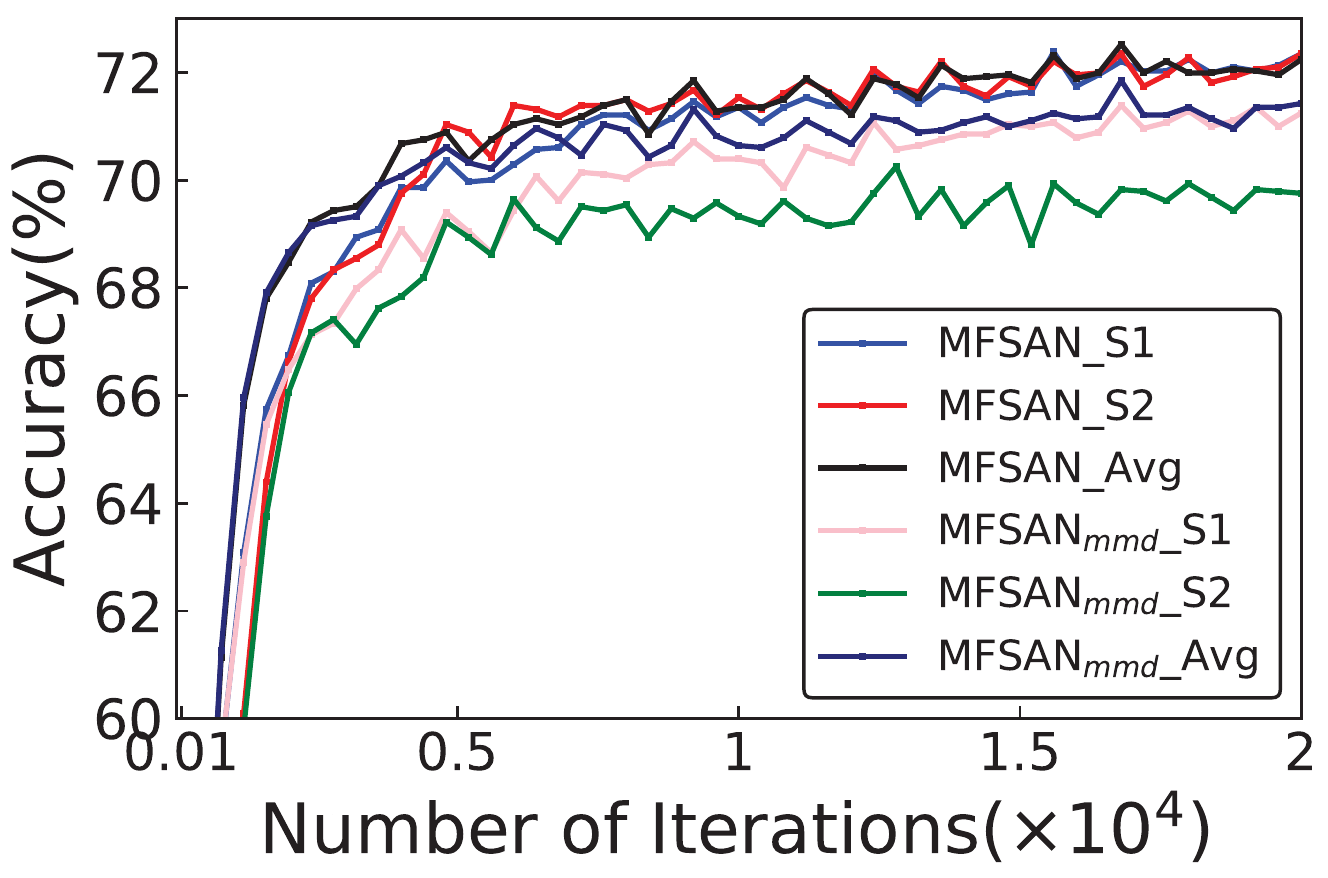}
		\label{fig:5b}
	}
	\subfigure[Accuracy w.r.t $\lambda$]{
		\centering
		\includegraphics[width=.47\columnwidth]{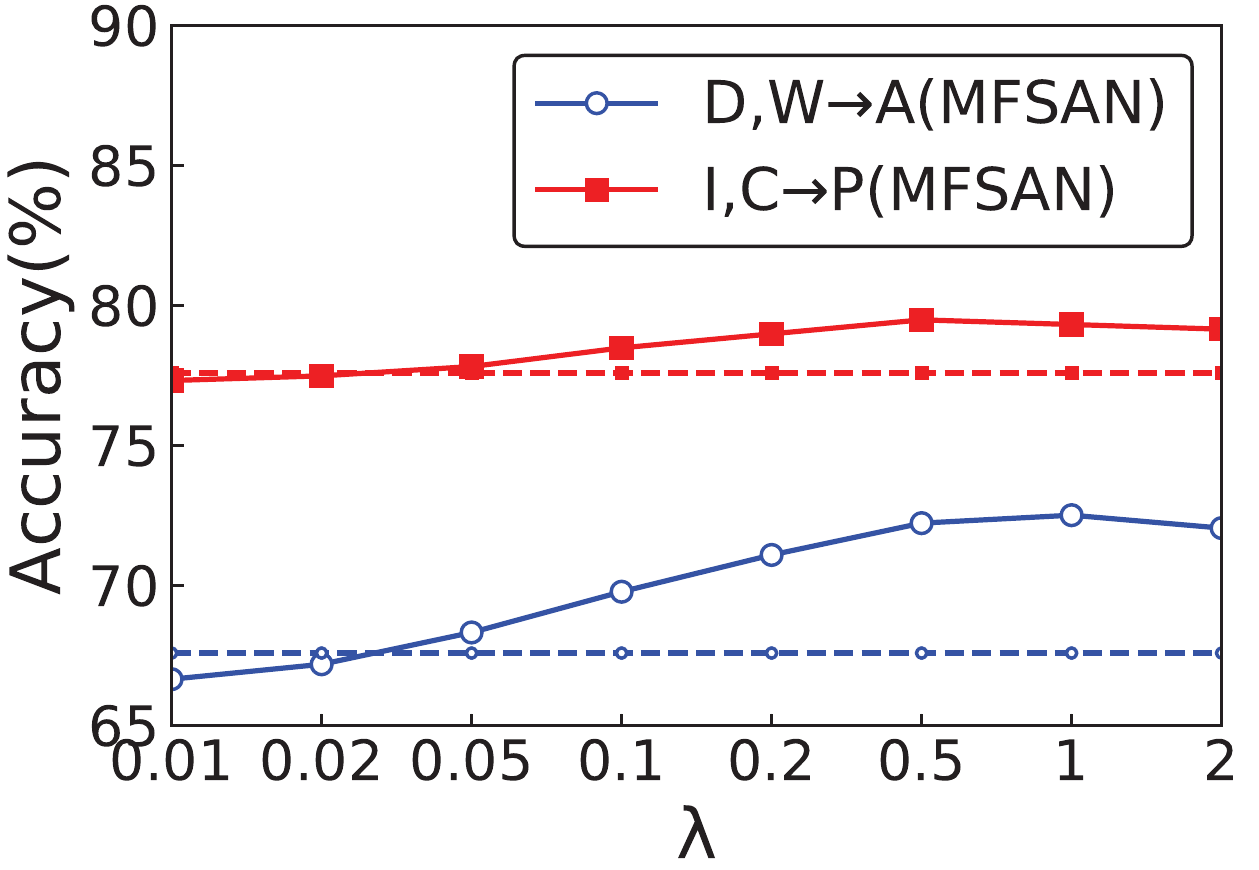}
		\label{fig:5c}
	}
	\caption{Algorithm Convergence and Parameter Sensitivity. }
	\label{fig:5}
\end{figure}

\textbf{Implementation Details} All deep methods are implemented base on the pytorch framework, and fine-tuned from pytorch-provided models of ResNet~\cite{he2016deep}. We fine-tune all convolutional and pooling layers and train the classifier layer via back propagation. Since the domain-specific feature extractors and classifiers are trained from scratch, we set its learning rate to be 10 times that of the other layers. We use mini-batch stochastic gradient descent (SGD) with momentum of 0.9 and the learning rate annealing strategy in RevGrad~\cite{ganin2015unsupervised}: the learning rate is not selected by a grid search due to high computational cost, it is adjusted during SGD using the following formula: ${\eta}_p = \frac{\eta_0}{(1+\alpha p)^\beta}$, where $p$ is the training progress linearly changing from $0$ to $1$, $\eta_0 = 0.01$, $\alpha = 10$ and $\beta = 0.75$, which is optimized to promote convergence and low error on the source domain. To suppress noisy activations at the early stages of training, instead of fixing the adaptation factor $\lambda$ and $\gamma$, we gradually change them from $0$ to $1$ by a progressive schedule: $\gamma_p = \lambda_p = \frac{2}{exp(-\theta p)} - 1$, and $\theta = 10$ is fixed throughout the experiments~\cite{ganin2015unsupervised}. This progressive strategy significantly stabilizes parameter sensitivity and eases model selection for MFSAN.

\subsection{Results}
we compare MFSAN with the baselines on three datasets and the results are shown in Tables~\ref{tab:MDAoffice31}, ~\ref{tab:MDAImage-CLEF} and~\ref{tab:MDAOfficeHome}, respectively. 
We also compare the MFSAN with or without disc loss on Office-31 dataset and list the results of each classifier from different sources and the average voting in Table~\ref{tab:diffoffice31}. 
From these results, we have the following insightful observations: 

$\bullet$ The results of Source Combine are better than Single Best, which demonstrates that combining all source domains into single source domain is helpful in most transfer tasks. This may be due to the data enrichment.

$\bullet$ MFSAN outperforms all compared baseline methods on most multi-source transfer tasks. The encouraging results indicate that it is important to learn multiple domain-invariant representations for each pair of source and target domains together with considering domain-specific class boundary.

$\bullet$ Comparing MFSAN$_{mmd}$ with DAN (source combine), the only difference is that MFSAN$_{mmd}$ extracts multiple domain-invariant representations in multiple feature spaces, while DAN extracts common domain-invariant representations in a common feature space. MFSAN$_{mmd}$ is better than DAN (Source Combine), which shows that it is difficult to extract common domain-invariant representations for all domains.

$\bullet$ MFSAN$_{disc}$ outperforms all compared methods on most multi-source transfer tasks. This verifies that the consideration of the domain-specific class boundary to reduce the gap between all classifiers can help each classifier learn the knowledge from other classifiers.

$\bullet$ Comparing MFSAN with MFSAN$_{mmd}$ which does not have disc loss, we find that the results of classifiers from different sources with disc loss (MFSAN) are very close to each other, while there is a large gap between the results of classifiers without disc loss (MFSAN$_{mmd}$). The results demonstrate the effectiveness of introducing disc loss to reduce the gap between all classifiers.

\subsection{Analysis}

\subsubsection{Feature visualization} In Figure~\ref{fig:4}, we visualize the latent representations of the task \textbf{D $\rightarrow$ A} learned by DAN (Single Source) and \textbf{D, W $\rightarrow$ A} learned by DAN (Source Combine), MFSAN using t-SNE embeddings~\cite{donahue2014decaf}. 

From Figure~\ref{fig:4}, we can observe that 1) the results in Figures~\ref{fig:4a} and~\ref{fig:4b} are better than the one in Figure~\ref{fig:5a}, which show that we can benefit from the consideration of more source domains: the results in Figures~\ref{fig:4c} and~\ref{fig:4d} are better than the ones in Figure~\ref{fig:5a} $\sim$~\ref{fig:4b}, which again validates the effectiveness of our model to align both domain-specific distributions and classifiers. 

\subsubsection{Algorithm Convergence} 
To investigate the convergence our algorithm and the influence of disc loss, we record the performance of MFSAN and MFSAN$_{mmd}$ during the iterating on the task \textbf{D, W $\rightarrow$ A} in Figure~\ref{fig:5b}. We can find that all algorithms can almost converge after $1.5\times 10^4$ iterations. Also, the results from MFSAN with disc loss have a smaller gap among classifiers and they achieve higher accuracy.

\subsubsection{Parameter Sensitivity}
For simplicity, we set the trade-off parameters $\lambda$ and $\gamma$ as the same value in our experiments, which respectively control the importance of mmd loss and disc loss. To study the sensitivity of $\lambda$, we sample the values in \{0.01, 0.02, 0.05, 0.1, 0.2, 0.5, 1, 2\}, and perform the experiments on tasks \textbf{D, W $\rightarrow$ A} and \textbf{I, C $\rightarrow$ P}. All the results are shown in Figure~\ref{fig:5c}, and we find that the accuracy first increases and then decreases, and displays as a bell-shaped curve. Finally, we set $\lambda = 0.5$ to achieve good performance.

\section{Conclusion}
\label{sec:conclusion}
Most previous deep learning based multi-source domain adaptation methods focus on extracting common domain-invariant representations for all domains without considering domain-specific class boundary. In this paper, we proposed a
Multiple Feature Space Adaptation Network (MFSAN), which simultaneously aligns the 
domain-specific distribution of each pair of source and target domains by learning multiple domain-invariant representations and the outputs of classifiers from multiple sources. Extensive experiments are conducted on three image datasets to demonstrate the effectiveness of the proposed framework. Moreover, our model is a general framework, which can integrate different kinds of mmd loss and disc loss functions. 

\section{Acknowledgments}
\label{sec:ack}
The research work is supported by the National Key Research and Development Program of China under Grant No. 2018YFB1004300, the National Natural Science Foundation of China under Grant No.61773361, 61473273, 91546122, Guangdong provincial science and technology plan projects under Grant No. 2015 B010109005, the Project of Youth Innovation Promotion Association CAS under Grant No. 2017146. Dr.Deqing Wang was supported by the National Natural Science Foundation of China (71501003).

\bibliographystyle{aaai}

\begin{thebibliography}{}

\bibitem[\protect\citeauthoryear{Ben-David \bgroup et al\mbox.\egroup
  }{2010}]{ben2010theory}
Ben-David, S.; Blitzer, J.; Crammer, K.; Kulesza, A.; Pereira, F.; and Vaughan,
  J.~W.
\newblock 2010.
\newblock A theory of learning from different domains.
\newblock {\em Machine learning} 79(1-2):151--175.

\bibitem[\protect\citeauthoryear{Blitzer \bgroup et al\mbox.\egroup
  }{2008}]{blitzer2008learning}
Blitzer, J.; Crammer, K.; Kulesza, A.; Pereira, F.; and Wortman, J.
\newblock 2008.
\newblock Learning bounds for domain adaptation.
\newblock In {\em NIPS},  129--136.

\bibitem[\protect\citeauthoryear{Bousmalis \bgroup et al\mbox.\egroup
  }{2017}]{bousmalis2017unsupervised}
Bousmalis, K.; Silberman, N.; Dohan, D.; Erhan, D.; and Krishnan, D.
\newblock 2017.
\newblock Unsupervised pixel-level domain adaptation with generative
  adversarial networks.
\newblock In {\em CVPR}, volume~1, ~7.

\bibitem[\protect\citeauthoryear{Donahue \bgroup et al\mbox.\egroup
  }{2014}]{donahue2014decaf}
Donahue, J.; Jia, Y.; Vinyals, O.; Hoffman, J.; Zhang, N.; Tzeng, E.; and
  Darrell, T.
\newblock 2014.
\newblock Decaf: A deep convolutional activation feature for generic visual
  recognition.
\newblock In {\em ICML},  647--655.

\bibitem[\protect\citeauthoryear{Duan, Xu, and Tsang}{2012}]{duan2012domain2}
Duan, L.; Xu, D.; and Tsang, I. W.-H.
\newblock 2012.
\newblock Domain adaptation from multiple sources: A domain-dependent
  regularization approach.
\newblock {\em IEEE TNNLS} 23(3):504--518.

\bibitem[\protect\citeauthoryear{Fernando \bgroup et al\mbox.\egroup
  }{2013}]{fernando2013unsupervised}
Fernando, B.; Habrard, A.; Sebban, M.; and Tuytelaars, T.
\newblock 2013.
\newblock Unsupervised visual domain adaptation using subspace alignment.
\newblock In {\em ICCV},  2960--2967.

\bibitem[\protect\citeauthoryear{Ganin and
  Lempitsky}{2015}]{ganin2015unsupervised}
Ganin, Y., and Lempitsky, V.
\newblock 2015.
\newblock Unsupervised domain adaptation by backpropagation.
\newblock In {\em ICML},  1180--1189.

\bibitem[\protect\citeauthoryear{Ghifary \bgroup et al\mbox.\egroup
  }{2016}]{ghifary2016deep}
Ghifary, M.; Kleijn, W.~B.; Zhang, M.; Balduzzi, D.; and Li, W.
\newblock 2016.
\newblock Deep reconstruction-classification networks for unsupervised domain
  adaptation.
\newblock In {\em ECCV},  597--613.

\bibitem[\protect\citeauthoryear{Gong \bgroup et al\mbox.\egroup
  }{2012}]{gong2012geodesic}
Gong, B.; Shi, Y.; Sha, F.; and Grauman, K.
\newblock 2012.
\newblock Geodesic flow kernel for unsupervised domain adaptation.
\newblock In {\em CVPR},  2066--2073.

\bibitem[\protect\citeauthoryear{Gretton \bgroup et al\mbox.\egroup
  }{2012}]{gretton2012kernel}
Gretton, A.; Borgwardt, K.~M.; Rasch, M.~J.; Sch{\"o}lkopf, B.; and Smola, A.
\newblock 2012.
\newblock A kernel two-sample test.
\newblock {\em JMLR} 13(Mar):723--773.

\bibitem[\protect\citeauthoryear{He \bgroup et al\mbox.\egroup
  }{2016}]{he2016deep}
He, K.; Zhang, X.; Ren, S.; and Sun, J.
\newblock 2016.
\newblock Deep residual learning for image recognition.
\newblock In {\em CVPR},  770--778.

\bibitem[\protect\citeauthoryear{Hoffman \bgroup et al\mbox.\egroup
  }{2018}]{hoffman2017cycada}
Hoffman, J.; Tzeng, E.; Park, T.; Zhu, J.-Y.; Isola, P.; Saenko, K.; Efros,
  A.~A.; and Darrell, T.
\newblock 2018.
\newblock Cycada: Cycle-consistent adversarial domain adaptation.
\newblock In {\em ICML}.

\bibitem[\protect\citeauthoryear{Huang \bgroup et al\mbox.\egroup
  }{2007}]{huang2007correcting}
Huang, J.; Gretton, A.; Borgwardt, K.~M.; Sch{\"o}lkopf, B.; and Smola, A.~J.
\newblock 2007.
\newblock Correcting sample selection bias by unlabeled data.
\newblock In {\em NIPS},  601--608.

\bibitem[\protect\citeauthoryear{Jhuo \bgroup et al\mbox.\egroup
  }{2012}]{jhuo2012robust}
Jhuo, I.-H.; Liu, D.; Lee, D.; and Chang, S.-F.
\newblock 2012.
\newblock Robust visual domain adaptation with low-rank reconstruction.
\newblock In {\em CVPR},  2168--2175.

\bibitem[\protect\citeauthoryear{Jiang and Zhai}{2007}]{jiang2007instance}
Jiang, J., and Zhai, C.
\newblock 2007.
\newblock Instance weighting for domain adaptation in nlp.
\newblock In {\em ACL},  264--271.

\bibitem[\protect\citeauthoryear{Liu, Shao, and Fu}{2016}]{liu2016structure}
Liu, H.; Shao, M.; and Fu, Y.
\newblock 2016.
\newblock Structure-preserved multi-source domain adaptation.
\newblock In {\em ICDM},  1059--1064.

\bibitem[\protect\citeauthoryear{Long \bgroup et al\mbox.\egroup
  }{2015}]{long2015learning}
Long, M.; Cao, Y.; Wang, J.; and Jordan, M.
\newblock 2015.
\newblock Learning transferable features with deep adaptation networks.
\newblock In {\em ICML},  97--105.

\bibitem[\protect\citeauthoryear{Long \bgroup et al\mbox.\egroup
  }{2016}]{long2016unsupervised}
Long, M.; Zhu, H.; Wang, J.; and Jordan, M.~I.
\newblock 2016.
\newblock Unsupervised domain adaptation with residual transfer networks.
\newblock In {\em NIPS},  136--144.

\bibitem[\protect\citeauthoryear{Long \bgroup et al\mbox.\egroup
  }{2017}]{long2016deep}
Long, M.; Zhu, H.; Wang, J.; and Jordan, M.~I.
\newblock 2017.
\newblock Deep transfer learning with joint adaptation networks.
\newblock In {\em ICML},  2208--2217.

\bibitem[\protect\citeauthoryear{Mansour, Mohri, and
  Rostamizadeh}{2009}]{mansour2009domain}
Mansour, Y.; Mohri, M.; and Rostamizadeh, A.
\newblock 2009.
\newblock Domain adaptation with multiple sources.
\newblock In {\em NIPS},  1041--1048.

\bibitem[\protect\citeauthoryear{Pan and Yang}{2010}]{pan2010survey}
Pan, S.~J., and Yang, Q.
\newblock 2010.
\newblock A survey on transfer learning.
\newblock {\em IEEE TKDE} 22(10):1345--1359.

\bibitem[\protect\citeauthoryear{Pan \bgroup et al\mbox.\egroup
  }{2011}]{pan2011domain}
Pan, S.~J.; Tsang, I.~W.; Kwok, J.~T.; and Yang, Q.
\newblock 2011.
\newblock Domain adaptation via transfer component analysis.
\newblock {\em IEEE TNN} 22(2):199--210.

\bibitem[\protect\citeauthoryear{Quionero-Candela \bgroup et al\mbox.\egroup
  }{2009}]{quionero2009dataset}
Quionero-Candela, J.; Sugiyama, M.; Schwaighofer, A.; and Lawrence, N.~D.
\newblock 2009.
\newblock {\em Dataset shift in machine learning}.
\newblock The MIT Press.

\bibitem[\protect\citeauthoryear{Ren \bgroup et al\mbox.\egroup
  }{2015}]{ren2015faster}
Ren, S.; He, K.; Girshick, R.; and Sun, J.
\newblock 2015.
\newblock Faster r-cnn: Towards real-time object detection with region proposal
  networks.
\newblock In {\em NIPS},  91--99.

\bibitem[\protect\citeauthoryear{Saenko \bgroup et al\mbox.\egroup
  }{2010}]{saenko2010adapting}
Saenko, K.; Kulis, B.; Fritz, M.; and Darrell, T.
\newblock 2010.
\newblock Adapting visual category models to new domains.
\newblock In {\em ECCV},  213--226.

\bibitem[\protect\citeauthoryear{Saito \bgroup et al\mbox.\egroup
  }{2017}]{saito2017maximum}
Saito, K.; Watanabe, K.; Ushiku, Y.; and Harada, T.
\newblock 2017.
\newblock Maximum classifier discrepancy for unsupervised domain adaptation.
\newblock In {\em CVPR}.

\bibitem[\protect\citeauthoryear{Sun and Saenko}{2016}]{sun2016deep}
Sun, B., and Saenko, K.
\newblock 2016.
\newblock Deep coral: Correlation alignment for deep domain adaptation.
\newblock In {\em ECCV},  443--450.

\bibitem[\protect\citeauthoryear{Tzeng \bgroup et al\mbox.\egroup
  }{2014}]{tzeng2014deep}
Tzeng, E.; Hoffman, J.; Zhang, N.; Saenko, K.; and Darrell, T.
\newblock 2014.
\newblock Deep domain confusion: Maximizing for domain invariance.
\newblock {\em arXiv preprint arXiv:1412.3474}.

\bibitem[\protect\citeauthoryear{Tzeng \bgroup et al\mbox.\egroup
  }{2015}]{tzeng2015simultaneous}
Tzeng, E.; Hoffman, J.; Darrell, T.; and Saenko, K.
\newblock 2015.
\newblock Simultaneous deep transfer across domains and tasks.
\newblock In {\em ICCV},  4068--4076.

\bibitem[\protect\citeauthoryear{Tzeng \bgroup et al\mbox.\egroup
  }{2017}]{tzeng2017adversarial}
Tzeng, E.; Hoffman, J.; Saenko, K.; and Darrell, T.
\newblock 2017.
\newblock Adversarial discriminative domain adaptation.
\newblock In {\em CVPR}, volume~1, ~4.

\bibitem[\protect\citeauthoryear{Venkateswara \bgroup et al\mbox.\egroup
  }{2017}]{venkateswara2017deep}
Venkateswara, H.; Eusebio, J.; Chakraborty, S.; and Panchanathan, S.
\newblock 2017.
\newblock Deep hashing network for unsupervised domain adaptation.
\newblock {\em arXiv preprint arXiv:1706.07522}.

\bibitem[\protect\citeauthoryear{Xu \bgroup et al\mbox.\egroup
  }{2018}]{xu2018deep}
Xu, R.; Chen, Z.; Zuo, W.; Yan, J.; and Lin, L.
\newblock 2018.
\newblock Deep cocktail network: Multi-source unsupervised domain adaptation
  with category shift.
\newblock In {\em CVPR},  3964--3973.

\bibitem[\protect\citeauthoryear{Yang, Yan, and
  Hauptmann}{2007}]{yang2007cross}
Yang, J.; Yan, R.; and Hauptmann, A.~G.
\newblock 2007.
\newblock Cross-domain video concept detection using adaptive svms.
\newblock In {\em MM},  188--197.

\bibitem[\protect\citeauthoryear{Zhao \bgroup et al\mbox.\egroup
  }{2018}]{zhao2018multiple}
Zhao, H.; Zhang, S.; Wu, G.; Gordon, G.~J.; et~al.
\newblock 2018.
\newblock Multiple source domain adaptation with adversarial learning.
\newblock In {\em ICLR}.

\end{thebibliography}

\end{document}